\documentclass[10pt,twocolumn,letterpaper]{article}

\usepackage[pagenumbers]{cvpr} 
\usepackage{algorithm}
\usepackage{algpseudocode}
\usepackage{bm}
\usepackage{booktabs}
\usepackage{mathtools}

\usepackage[dvipsnames]{xcolor}

\let\oldciteauthor=\citeauthor
\def\citeauthor#1{\hypersetup{citecolor=black}\oldciteauthor{#1}}
\let\oldcite=\cite
\def\cite#1{\hypersetup{citecolor=cvprblue}\oldcite{#1}}

\definecolor{cvprblue}{rgb}{0.21,0.49,0.74}
\usepackage[pagebackref,breaklinks,colorlinks,citecolor=cvprblue]{hyperref}

\def\paperID{7394} 
\def\confName{CVPR}
\def\confYear{2024}

\title{MotionEditor: Editing Video Motion via Content-Aware Diffusion}

\author{Shuyuan Tu\textsuperscript{1,2}, Qi Dai\textsuperscript{3}, Zhi-Qi Cheng\textsuperscript{4}, Han Hu\textsuperscript{3}, Xintong Han\textsuperscript{5},
Zuxuan Wu\textsuperscript{1,2}, Yu-Gang Jiang\textsuperscript{1,2}\\
{\textsuperscript{1}Shanghai Key Lab of Intell. Info. Processing, School of CS, Fudan University} \\
{\textsuperscript{2}Shanghai Collaborative Innovation Center of Intelligent Visual Computing} \\
{\textsuperscript{3}Microsoft Research Asia ~~ \textsuperscript{4}Carnegie Mellon University ~~ \textsuperscript{5}Huya Inc.} \\
{\url{https://francis-rings.github.io/MotionEditor/}}
}

\begin{document}
\twocolumn[{
\maketitle
\vspace{-1.4em}
\renewcommand\twocolumn[1][]{#1}
\begin{center}
    \centering
    \includegraphics[width=1\textwidth]{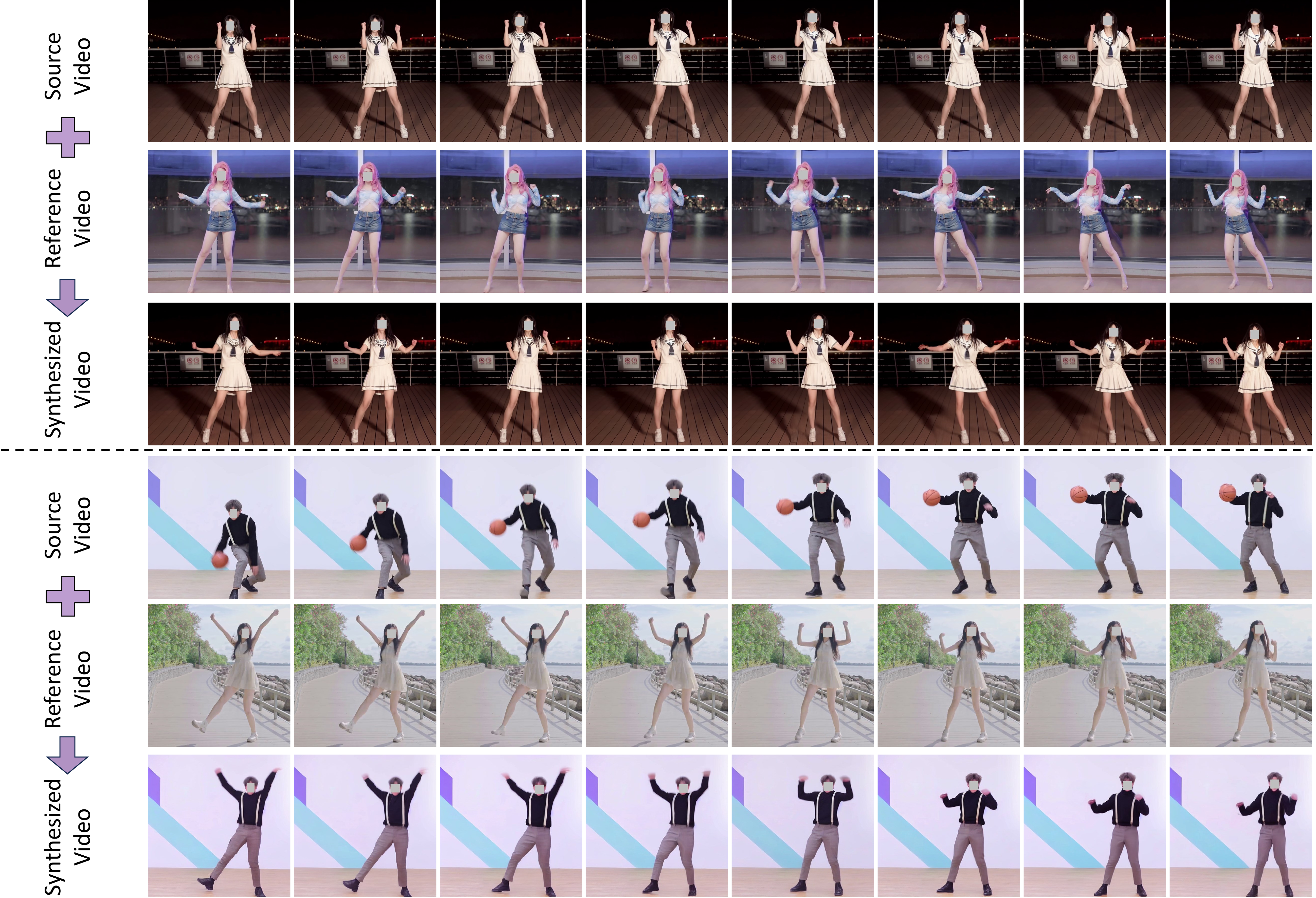}
    \vspace{-0.5cm}
    \captionof{figure}{
    MotionEditor: A diffusion-based video editing method aimed at transferring motion from a reference to a source.
    }
    \label{fig:editing_result}
    \vspace{-0.0cm}
\end{center}
}]

\begin{abstract}
Existing diffusion-based video editing models have made gorgeous advances for editing attributes of a source video over time but struggle to manipulate the motion information while preserving the original protagonist's appearance and background.
To address this, we propose MotionEditor, a diffusion model for video motion editing.
MotionEditor incorporates a novel content-aware motion adapter into ControlNet to capture temporal motion correspondence.
While ControlNet enables direct generation based on skeleton poses, it encounters challenges when modifying the source motion in the inverted noise due to contradictory signals between the noise (source) and the condition (reference).
Our adapter complements ControlNet by involving source content to transfer adapted control signals seamlessly.
Further, we build up a two-branch architecture (a reconstruction branch and an editing branch) with a high-fidelity attention injection mechanism facilitating branch interaction. 
This mechanism enables the editing branch to query the key and value from the reconstruction branch in a decoupled manner, making the editing branch retain the original background and protagonist appearance.
We also propose a skeleton alignment algorithm to address the discrepancies in pose size and position.
Experiments demonstrate the promising motion editing ability of MotionEditor, both qualitatively and quantitatively.
\end{abstract}

\section{Introduction}
\label{sec:intro}
Diffusion models~\cite{dhariwal2021diffusion,ho2020denoising,ho2022cascaded,nichol2021improved,song2020score,song2020denoising,rombach2022high,meng2021sdedit,hertz2022prompt,tumanyan2023plug} have achieved remarkable success in image generation, which inspired plenty studies in video editing~\cite{wu2023tune,liu2023video,qi2023fatezero,ceylan2023pix2video,khachatryan2023text2video}.
While significant progress has been made, existing diffusion models for video editing primarily focus on texture editing, such as attribute manipulation for protagonists, background editing, and style editing. 
The motion information, which stands out as the most unique and distinct feature when compared to images, is mostly ignored.
This raises the question: Can we manipulate the motion of a video in alignment with a reference video?
In this paper, we attempt to explore a novel, higher-level, and more challenging video editing scenario---motion editing. 
Given a reference video and prompt, we aim to change the protagonist's motion in the source to match the reference video while preserving the original appearance.

To date in literature, researchers have explored human motion transfer~\cite{liu2019liquid,siarohin2019first,siarohin2021motion} and pose-guided video generation~\cite{ma2023follow,zhang2023controlvideo}. The former focuses on animating still images based on reference skeletons, while the latter tries to generate pose-aligned videos without preserving a desired appearance.
Motion editing differs in that it directly modifies the motion in the video while preserving other video information, such as dynamic per-frame background and camera movements.

Recent advancements in visual editing have mainly emerged through the use of diffusion models~\cite{cao2023masactrl, shi2023dragdiffusion, mou2023dragondiffusion,zhang2023adding,ma2023follow,zhang2023controlvideo}.
For example, ControlNet~\cite{zhang2023adding} enables direct controllable generation conditioned on poses.
However, they suffer from severe artifacts when trying to edit the motion according to another pose.
We hypothesize the reason is that the control signals are only derived from the reference poses, which cannot properly accommodate the source content, thus resulting in a contradiction.
Some methods also lack the ability to preserve the appearance of the protagonist and background, as well as the temporal consistency. In this paper, we rethink the fundamental efficacy of ControlNet.
We argue that it is essential to involve the source latent when generating the controlling signal for the editing task. With the help of source contents, the condition guidance can precisely perceive the entire context and structures, and adjust its distribution to prevent undesired distortion.

To this end, we propose MotionEditor, depicted in Fig. \ref{fig:overview}, to take one step forward in exploring video motion editing with diffusion models. 
MotionEditor requires one-shot learning on a source video to preserve the original texture feature. 
We then introduce a content-aware motion adapter appended to ControlNet for enhancing control capability and temporal modeling.
The adapter consists of content-aware blocks and temporal blocks.
In particular, content-aware blocks perform cross-attention to incorporate the source frame feature, which significantly boosts the quality of motion control.

At inference time, a skeleton alignment algorithm is devised to counter the size and position disparities between source skeletons and reference skeletons. We further propose an attention injection mechanism based on a two-branch architecture (reconstruction and editing branches) to preserve the source appearance of the protagonist and the background.
Previous attention fusion strategies directly inject the attention map or key into the editing branch. The direct injection may result in confusion between the edited foreground and background. 
In some cases, it would bring noise to the editing branch, thereby resulting in the phenomenon of overlapping and shadow flickering. 
To avoid this, we propose to decouple the keys/values in the foreground and background using a segmentation mask.
The keys/values in the editing branch are thus supplemented by those from the foreground and background in the reconstruction branch.
As such, the editing branch is able to capture the background details and the geometric structure of the protagonist from the source.

In conclusion, our contributions are as follows:
(1)  We explore the video diffusion models for motion editing, which is usually ignored by previous video editing works.
(2) A novel content-aware motion adapter is proposed to enable the ControlNet to perform consistent and precise motion editing. 
(3) We propose a high-fidelity attention injection mechanism that preserves the background information and the geometric structure of the protagonist from the source. The mechanism is only active during inference, making it training-free.
(4) We conduct experiments on in-the-wild videos, where the results show the superiority of our method compared with the state-of-the-art.

\section{Related Work}
\label{sec:related_work}
\noindent\textbf{Diffusion for Image Editing} ~Image generation has been remarkably improved with diffusion models~\cite{rombach2022high}, surpassing previous GAN models~\cite{goodfellow2020generative, gal2022stylegan, park2019semantic, patashnik2021styleclip, wang2018high} in quality and editing capabilities.
Building on the pioneering work by \cite{rombach2022high}, researchers have made progress in this direction.
\citeauthor{meng2021sdedit} propose SDEdit, the first approach that enables image editing via inversion and reversion. This allows more precise control over the edits compared with previous GAN-based approaches.
Prompt-to-Prompt~\cite{hertz2022prompt} and Plug-and-Play~\cite{tumanyan2023plug} introduce techniques to generate more coherent hybrid images, tackling the inconsistent noise.
Meanwhile, UniTune~\cite{valevski2022unitune} and Imagic~\cite{kawar2023imagic} focus on personalized fine-tuning of diffusion models. In short, these approaches enable edits that remain truer to the original image.

To generate diverse contents, researchers have proposed several controllable diffusion models~\cite{zhang2023adding, huang2023composer, saharia2022palette, gal2022image, ruiz2023dreambooth, nikankin2022sinfusion}. \citeauthor{zhang2023adding} introduce ControlNet, which enables Stable Diffusion~\cite{rombach2022high} to embrace multiple controllable conditions for text-to-image synthesis.
\citeauthor{cao2023masactrl} propose MasaCtrl, which converts existing self-attention into mutual self-attention to achieve non-rigid image synthesis. 
Overall, these approaches have advanced diffusion-based image editing through the handling of inconsistent noise, fine-tuning strategies, and inversion-based editing. However, it still remains an open challenge to enable precise semantic edits while maintaining perfect fidelity and coherence.

\vspace{0.1cm}
\noindent\textbf{Diffusion Models for Video Editing and Generation} ~Video editing is complex that demands temporal consistency in editing. 
Most methods use existing text-to-image diffusion models with an additional temporal modeling module. 
Tune-a-Video~\cite{wu2023tune} and Text2Video-Zero~\cite{khachatryan2023text2video} inflate 2D diffusion models to 3D models.
FateZero~\cite{qi2023fatezero} and Vid2Vid-zero~\cite{wang2023zero} use mutual attention to ensure consistency of geometry and color information.
On the other hand, Text2LIVE~\cite{bar2022text2live} and StableVideo~\cite{chai2023stablevideo} decompose video semantics by leveraging layered neural atlas~\cite{kasten2021layered}. However, these models are limited to low-level attribute editing and cannot edit complex information such as motion.

Recently, pose-driven video generation has become popular.
Follow-Your-Pose~\cite{ma2023follow} extracts coarse temporal motion information by implementing an adapter on the text-to-image diffusion model. ControlVideo~\cite{zhang2023controlvideo} introduces full cross-frame attention by extending ControlNet to videos. However, these methods focus on video generation rather than motion editing, which can result in the distortion of the protagonist and background. Unlike these models, our MotionEditor aims to perform motion editing while preserving the appearance of the original video.

\vspace{0.1cm}
\noindent\textbf{Human Motion Transfer} ~This task aims to transfer the motion from a video to a target image, enabling animation of the target. Previous GAN methods \cite{liu2019liquid,siarohin2019first,siarohin2021motion} have tackled this task but struggle with complex motions and backgrounds.
LWG \cite{liu2019liquid} uses 3D pose projection for motion transfer but cannot model internal motions well.
FOMM~\cite{siarohin2019first} approximates motion transfer via boundary key points.
MRAA~\cite{siarohin2021motion} exploits regional features to capture part motion, yet the performance is limited on complex scenes.
To address these limitations, we propose MotionEditor for high-quality motion editing on complex motions and backgrounds. In contrast to prior work, it leverages diffusion models capable of \emph{generating consistent details} even for intricate motions and backgrounds.



\section{Preliminaries}
\label{sec:preliminaries}

\begin{figure*}[t!]
\begin{center}
\includegraphics[width=1\linewidth]{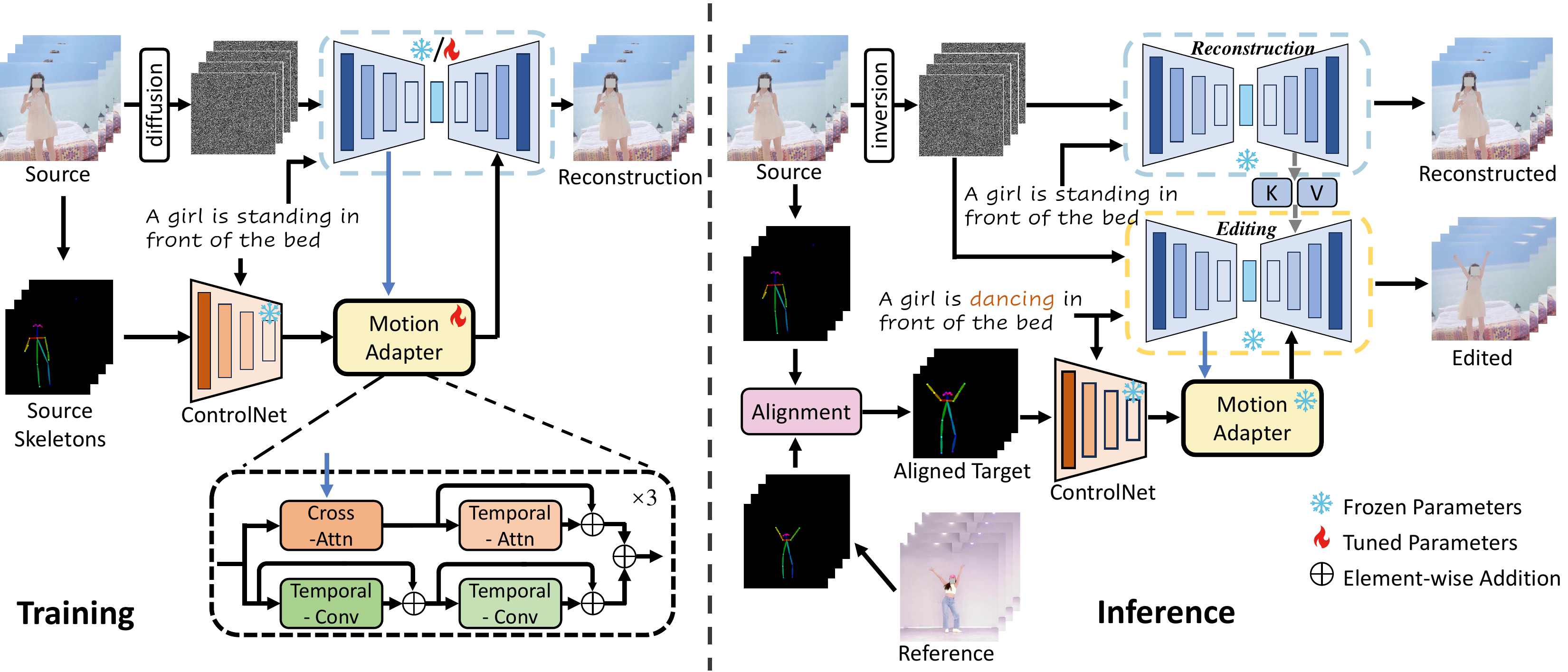}
\end{center}
\vspace{-0.5cm}
   \caption{Architecture overview of MotionEditor. In training, only the motion adapter and temporal attention in U-Net are trainable. 
   In inference, we first align the source and reference skeletons through resizing and translation.
   We then build a two-branch framework: one for reconstruction and the other for editing.
   Motion adapter enhances the motion guidance of ControlNet by leveraging the information from the source latent.
   We also inject the key/value in the reconstruction branch into the editing branch to preserve the source appearance.
   }
\label{fig:overview}
\vspace{-0.1cm}
\end{figure*}

Diffusion models~\cite{ho2020denoising, nichol2021improved, song2020denoising} have recently shown promising results for
synthesizing high-quality and diverse contents using iterative denoising operations. They consist of a forward diffusion process and a reverse denoising process. During the forward process, models equipped with a predefined noise schedule $\alpha_t$ add random noise to the source sample $\bm{x}_0$ at time step $t$ for obtaining a noised sample $\bm{x}_t$:
\begin{equation}\small
\label{eq:diffusion_forward}
\begin{aligned}
     &\bm{q}(\bm{x}_{1:T}) = \bm{q}(\bm{x}_{0})\prod_{t=1}^{T}\bm{q}(\bm{x}_{t}|\bm{x}_{t-1}), \\
     &\bm{q}(\bm{x}_{t}|\bm{x}_{t-1}) = \mathcal{N}(\bm{x}_{t}; \sqrt{\alpha_{t}}\bm{x}_{t-1},(1-\alpha_{t})\mathbf{I}).
\end{aligned}
\end{equation}
The original input $\bm{x}_{0}$ is inverted into Gaussian noise $\bm{x}_{T}\sim \mathcal{N}(\mathbf{0}, \mathbf{1})$ after $T$ forward steps. The reverse process attempts to predict a cleaner sample $\bm{x}_{t-1}$ based on $\bm{x}_{t}$ by removing noise.
The process is depicted as:
\begin{equation}\small
\label{eq:diffusion_reversion}
\begin{aligned}
     \bm{p}_{\theta}(\bm{x}_{t-1}|\bm{x}_{t}) = \mathcal{N}(\bm{x}_{t-1}; \bm{\mu}_{\theta}(\bm{x}_{t}, t), \bm{\sigma}_{t}^{2}\mathbf{I}),
\end{aligned}
\end{equation}
where $\bm{\mu}_{\theta}(\bm{x}_{t}, t)$ and $\bm{\sigma}_{t}^2$ indicate the mean and variance of the sample at the current time step. Only the mean is correlated with the time step and noise, while the variance is constant. The denoising network $\bm{\varepsilon}_{\theta}(\bm{x}_{t},t)$ aims to predict the noise $\bm{\varepsilon}$ by training with a simplified mean squared error:
\begin{equation}\small
\label{eq:diffusion_loss}
\begin{aligned}
     \mathcal{L}_{simple} = \mathbb{E}_{\bm{x}_{0},\bm{\varepsilon},t}(\left \| \bm{\varepsilon} -\bm{\varepsilon}_{\theta}(\bm{x}_{t}, t)  \right \|^{2}).
\end{aligned}
\end{equation}
Once the model is trained, we feed $\bm{x}_{T}\sim \mathcal{N}(\mathbf{0}, \mathbf{1})$ to the diffusion model and iteratively perform DDIM sampling for predicting cleaner $\bm{x}_{t-1}$ from the noise sample $\bm{x}_{t}$ of a previous time step. The process is shown as follows:
\begin{equation}\small
\label{eq:diffusion_loss}
\begin{aligned}
     \bm{x}_{t-1} = \sqrt{\alpha_{t-1}}\frac{\bm{x}_{t}-\sqrt{1-\alpha_{t}}\bm{\varepsilon}_{\theta}(\bm{x}_{t}, t)}{\sqrt{\alpha_{t}}}+\sqrt{1-\alpha_{t-1}}\bm{\varepsilon}_{\theta}(\bm{x}_{t}, t).
\end{aligned}
\end{equation}
One can also inject a text prompt $\bm{p}$ into the prediction model $\bm{\varepsilon}_{\theta}(\bm{x}_{t}, t, \bm{p})$ as 
 a condition, where the diffusion model can perform T2I synthesis. 
Recent work~\cite{rombach2022high} introduces an encoder $\mathcal{E}$ to compress images $\bm{x}$ into a latent space $\bm{z}=\mathcal{E}(\bm{x})$, and a decoder $\mathcal{D}$ to transfer the latent embedding back to the pixel space. In this way, the diffusion process is performed in the latent space.

\section{Method}
\label{sec:method}

\subsection{Architecture Overview}
\label{sec: architecture_overview}

As illustrated in Fig. \ref{fig:overview}, MotionEditor is based on the commonly used T2I diffusion model (\emph{i.e.} LDM~\cite{rombach2022high}) and ControlNet~\cite{zhang2023adding}.
We first inflate the spatial transformer in the U-Net of the LDM to a 3D transformer by appending a temporal self-attention layer.
We also propose Consistent-Sparse attention, described in Sec.~\ref{sec: attention_fusion}, to replace the spatial self-attention.
To achieve precise motion manipulation and temporal consistency, we design a content-aware motion adapter, operating on features from the U-Net and conditional pose information from the ControlNet. Following~\cite{wu2023tune}, we perform one-shot training to compute the weights for the temporal attention module and the motion adapter to reconstruct the input source video.

During inference, given a reference target video $\bm{X}_{rf}$, MotionEditor aims to transfer the motion of $\bm{X}_{rf}$ to the source while preserving the appearance information of the source.
To this end, we first develop a skeleton alignment algorithm to narrow the gap between source skeleton $\bm{S}_{sr}$ and reference skeleton $\bm{S}_{rf}$ by considering the position and size, and 
produce a refined target skeleton $\bar{\bm{S}}_{tg}$.
We then employ the DDIM inversion~\cite{song2020denoising} on pixel values of the video to produce a latent noise that serves as the starting point for sampling.
More importantly, we introduce a high-fidelity attention injection module exploring a carefully designed two-branch network. One is dedicated to reconstruction, while the other one focuses on editing. More specifically, the editing branch takes features from the refined target skeleton as inputs, transferring the motion information from the reference to the source. Meanwhile, critical appearance information encoded in the reconstruction branch is further injected into the editing branch so that the appearance and the background are preserved. Below, we introduce the proposed components in detail.

\subsection{Content-Aware Motion Adapter}
\label{sec: adapter}
Our goal is to manipulate the body motion in videos under the guidance of pose signals.
While ControlNet~\cite{zhang2023adding} enables direct controllable generation based on conditions, it has difficulties modifying the source motion from the inverted noise.
The motion signals injected by the ControlNet could conflict with the source motion, thereby resulting in pronounced ghosting and blurring effects or even the loss of controlling ability.
Furthermore, the model is derived from an image, lacking the ability to generate temporal consistent contents.
Therefore, we propose a temporal content-aware motion adapter that enhances motion guidance as well as facilitates temporal consistency, as shown in Fig. \ref{fig:overview}.

Our motion adapter takes as input the feature output by ControlNet, which has been observed to achieve promising spatial control.
Instead of inserting temporal layers into ControlNet, we leave it as is to prevent prejudicing its inherent modeling capability.
The adapter consists of two parallel paths corresponding to different perception granularity.
One is the global modeling path, including a content-aware cross-attention block and a temporal attention block. The other is the local modeling path, which uses two temporal convolution blocks to capture local motion features.
Specifically, our cross-attention involves the latent feature from the U-Net to model the pose feature, where the query comes from the pose feature $\bm{m}_{i}$, and the key/value is from the corresponding frame latent $\bm{z}_{i}$ produced by the U-Net:
\begin{equation}\small
\label{eq:overview_cross_attn}
\begin{aligned}
     \bm{Q}=\bm{W}_c^{Q}\bm{m}_{i},\bm{K}=\bm{W}_c^{K}\bm{z}_{i},\bm{V}=\bm{W}_c^{V}\bm{z}_{i},
\end{aligned}
\end{equation}
where {\small$\bm{W}_c^{Q}$, $\bm{W}_c^{K}$} and {\small$\bm{W}_c^{V}$} are projection matrices.
Our cross-attention enables the motion adapter to concentrate on correlated motion clues in the video latent space, which significantly enhances the control capability.
By building up a bridge between them, the model can thereby manipulate the source motion seamlessly, without contradiction.

\subsection{High-Fidelity Attention Injection}
\label{sec: attention_fusion}

\begin{figure}[t!]
\begin{center}
\includegraphics[width=1\linewidth]{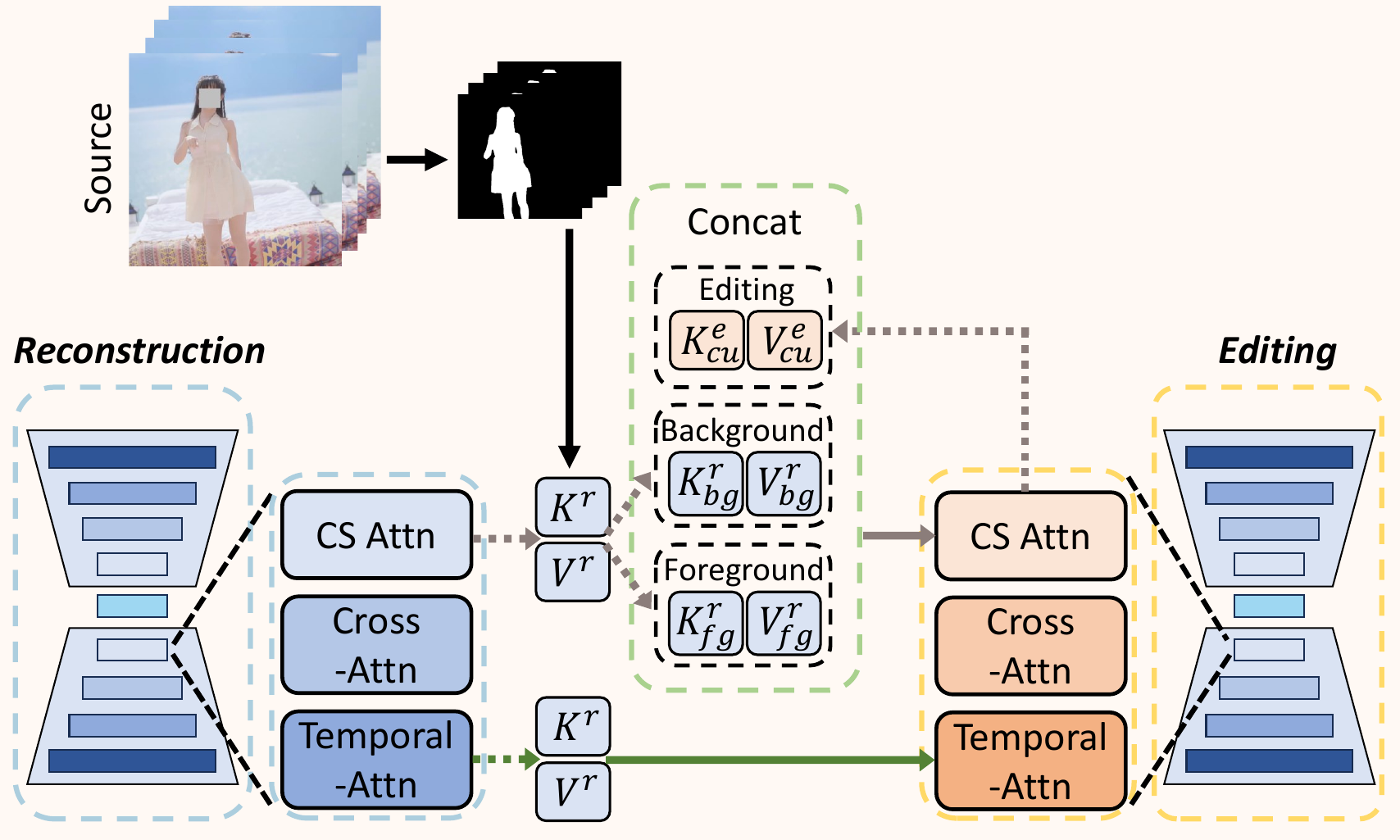}
\end{center}
\vspace{-0.6cm}
   \caption{Illustration of high-fidelity attention injection during inference. We leverage the source foreground masks to guide the decoupling of key/value in the Consistent-Sparse Attention.}
\label{fig:attention}
\vspace{-0.4cm}
\end{figure}

Although our motion adapter can accurately capture body poses, it may undesirably alter the appearance of the protagonist and the background.
Consequently, we propose a high-fidelity attention injection from the reconstruction branch to the editing branch, which preserves the details of the subject and background in the synthesized video.
While previous attention fusion paradigms~\cite{qi2023fatezero,wang2023zero,cao2023masactrl} have used attention maps or keys/values for editing, they suffer from severe quality degradation in ambiguous regions, \ie motion areas, due to context confusion.
To solve the issue, we decouple the keys and values into foreground and background through semantic masks.
By injecting separated keys and values into the editing branch from the reconstruction branch, we can reduce the confusion hindering the editing.
The pipeline is shown in Fig. \ref{fig:attention}.

Before introducing the injection, we first present the details of attention blocks in the model.
Each attention block in U-Net consists of our designed Consistent-Sparse Attention (CS Attention), Cross Attention, and Temporal Attention. 
The Consistent-Sparse Attention, as a sparse causal attention, replaces the spatial attention in the original U-Net.
It aims to perform spatiotemporal modeling with little additional computational overhead.
Specifically, taking the reconstruction branch as an example, the query in CS Attention is derived from the current frame $\bm{z}^r_{i}$, while the key/value is obtained from the preceding and current frames $\bm{z}^r_{i-1},\bm{z}^r_{i}$. 
This design can improve the frame consistency:
\begin{equation}\small
\label{eq:attention_sparse}
\begin{aligned}
     \bm{Q}^r=\bm{W}^{Q}\bm{z}^r_{i},\bm{K}^r=\bm{W}^{K}[\bm{z}^r_{i-1}, \bm{z}^r_{i}],\bm{V}^r=\bm{W}^{V}[\bm{z}^r_{i-1}, \bm{z}^r_{i}],
\end{aligned}
\end{equation}
where $[\cdot]$ refers to concatenation. {\small$\bm{W}^{Q}$, $\bm{W}^{K}$} and {\small$\bm{W}^{V}$} are projection matrices.
It is worth noting that we do not employ the sparse attention in~\cite{wu2023tune}, in which the key/value is from the first and preceding frames $\bm{z}_{0},\bm{z}_{i-1}$.
We find that it may force the synthesized motion to favor the first frame excessively, resulting in flickering.

We now present the injection of keys and values from the reconstruction branch to those of the editing branch, operating on both Consistent-Sparse Attention (CS Attention) and Temporal Attention. The injection is only active in the decoder of U-Net. For CS Attention, we leverage a source foreground mask {\small$\bm{M}$}, obtained from an off-the-shelf segmentation model, to decouple the foreground and background information.
Given the key {\small$\bm{K}^r$} and value {\small$\bm{V}^r$} in the reconstruction branch, we separate them into the foreground ({\small$\bm{K}^{r}_{fg}$} and {\small$\bm{V}^{r}_{fg}$}) and background ({\small$\bm{K}^{r}_{bg}$} and {\small$\bm{V}^{r}_{bg}$}),
\begin{equation}\small
\label{eq:attention_decouple}
\begin{aligned}
     \bm{K}^r_{fg}&=\bm{K}^r\odot \bm{M}, ~~\bm{V}^r_{fg}=\bm{V}^{r}\odot \bm{M}, \\
     \bm{K}^{r}_{bg}&=\bm{K}^{r}\odot (\bm{1}-\bm{M}), ~~\bm{V}^{r}_{bg}=\bm{V}^{r}\odot (\bm{1}-\bm{M}).
\end{aligned}
\end{equation}
The decoupling operation introduces an explicit distinction between background and foreground. It encourages the model to focus more on the individual appearance instead of mixing both, ensuring the high fidelity of the subject and background.
Notably, simply replacing the key/value in the editing branch with the above ones would cause a large number of abrupt motion changes, as the model is significantly influenced by the source motion.
Instead, we combine them to maintain the target motion precisely.
Therefore, in the editing branch, the key and value of CS Attention are updated by the injected key {\small$\bm{K}_{inj}$} and value {\small$\bm{V}_{inj}$}:
\begin{equation}\small
\label{eq:attention_decouple}
\begin{aligned}
     &\bm{K}^{e} = \bm{W}^{K}[\bm{z}^e_{i-1},\bm{z}^e_{i}] = [\bm{W}^{K}\bm{z}^e_{i-1},\bm{W}^{K}\bm{z}^e_{i}]\mathrel{\coloneqq} [\bm{K}^{e}_{pr},\bm{K}^{e}_{cu}], \\
     &\bm{V}^{e} = \bm{W}^{V}[\bm{z}^e_{i-1},\bm{z}^e_{i}] = [\bm{W}^{V}\bm{z}^e_{i-1},\bm{W}^{V}\bm{z}^e_{i}]\mathrel{\coloneqq} [\bm{V}^{e}_{pr},\bm{V}^{e}_{cu}], \\
     &\bm{K}_{inj}=[\bm{K}^{r}_{fg}, \bm{K}^{r}_{bg}, \bm{K}^{e}_{cu}], \\
     &\bm{V}_{inj}=[\bm{V}^{r}_{fg}, \bm{V}^{r}_{bg}, \bm{V}^{e}_{cu}],
\end{aligned}
\end{equation}
where $\bm{z}^e_{i},\bm{z}^e_{i-1}$ indicate the current and preceding frames in the editing branch. {\small$\bm{K}^{e},\bm{V}^{e}$} are the original key and value in CS attention of the editing branch. {\small$\bm{K}^{e}_{cu},\bm{V}^{e}_{cu}$} are the key and value from the current frame.

The injection in temporal attention is much simpler than in CS Attention since it already performs the local operation concerning the spatial region.
We directly inject the {\small$\bm{K}^{r},\bm{V}^{r}$} of reconstruction branch into the editing branch.

\begin{figure*}[t!]
\begin{center}
\includegraphics[width=0.98\linewidth]{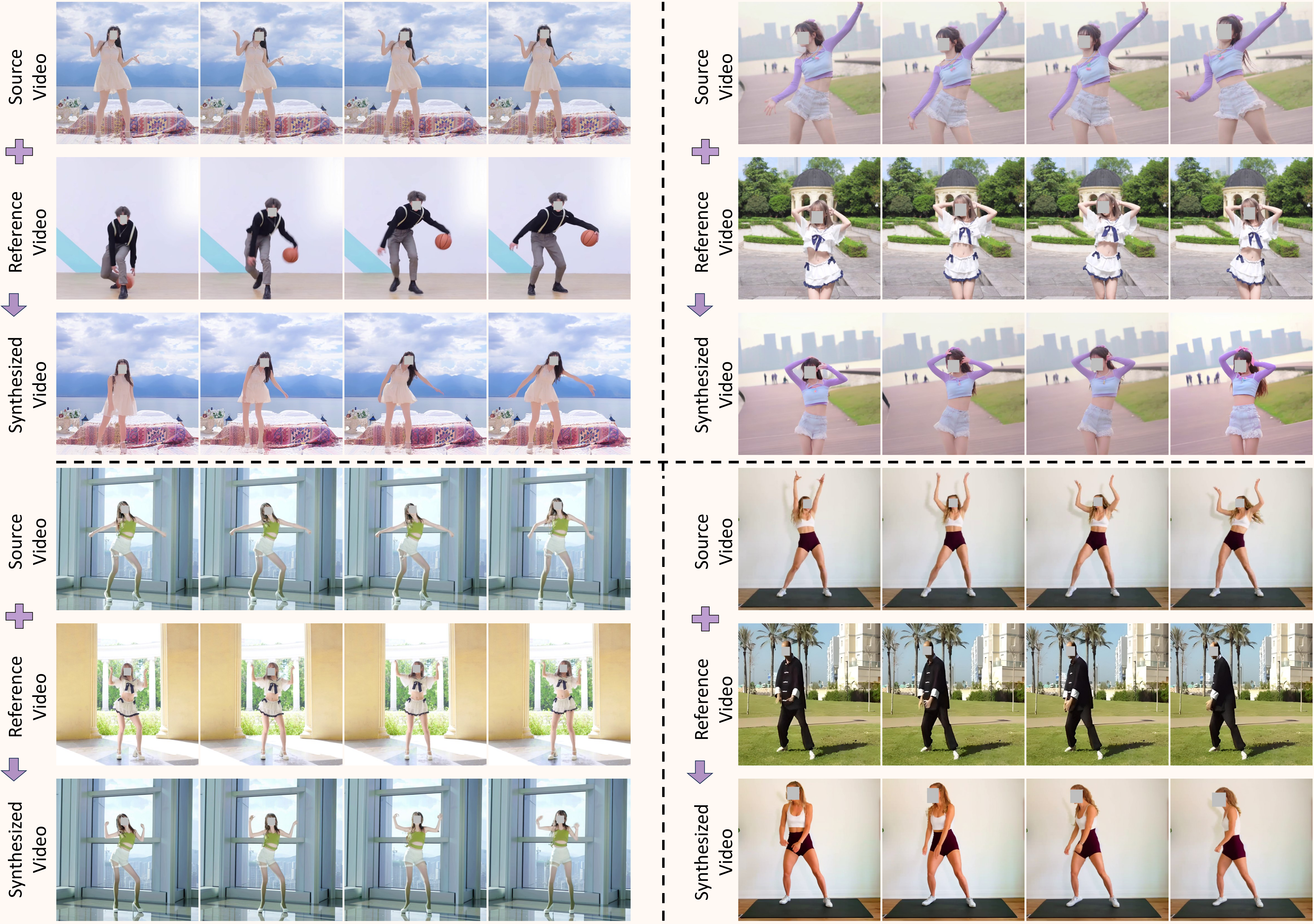}
\end{center}
\vspace{-0.55cm}
   \caption{Motion editing results of our MotionEditor. More examples can be found in the appendix.}
\label{fig:editing_result}
\vspace{-0.4cm}
\end{figure*}

\subsection{Skeleton Signal Alignment}
\label{sec: skeleton_alignment}
Given the source and reference videos, there always exists a gap between the source and the target protagonists due to different sizes and coordinated positions. The discrepancy may affect the performance of editing. Therefore, we propose an alignment algorithm for addressing this issue. 

Our algorithm contains two steps, namely a resizing operation and a translation operation.
Concretely, given the source video $\bm{X}_{sr}$ and a reference video $\bm{X}_{rf}$, we first extract the source skeleton $\bm{S}_{sr}$ and a foreground mask $\bm{M}_{sr}$, as well as the reference skeleton $\bm{S}_{rf}$ and its mask $\bm{M}_{rf}$, using off-the-shelf models.
We then perform edge detection on the masks to obtain rectangular outlines of the foreground.
Based on the area of two rectangular outlines, we scale $\bm{S}_{rf}$ to the same size as the source.
Regarding the foreground position, we calculate the average coordinate of the foreground pixels in each mask, which denotes the center of the protagonist.
An offset vector is computed by calculating the difference between the two centers.
With this vector, we obtain an affine matrix for the translation operation, which is further applied to the resized reference skeleton.
Finally, the target skeleton $\bar{\bm{S}}_{tg}$ is generated.
The details of alignment are depicted in the appendix.

\section{Experiments}
\label{sec:method}

\subsection{Implementation Details}
Our proposed MotionEditor is based on the Latent Diffusion Model~\cite{rombach2022high} (Stable Diffusion). We evaluate our model on YouTube videos and videos from the TaichiHD~\cite{siarohin2019first} dataset, in which each video includes a protagonist of at least 70 frames. The resolution of frames is unified to $512\times 512$. We perform one-shot learning for the motion adapter for 300 steps with a constant learning rate of $3\times 10^{-5}$. We employ DDIM inversion~\cite{song2020denoising} and null-text optimization~\cite{mokady2023null} with classifier-free guidance~\cite{ho2022classifier} during inference. 
Due to the usage of DDIM inversion and null-text optimization, MotionEditor needs 10 minutes to perform motion editing for each video on a single NVIDIA A100 GPU.

\subsection{Motion Editing Results}

We validate the motion editing superiority of our proposed MotionEditor extensively. Here, we demonstrate several cases in Figure \ref{fig:editing_result} and the rest of the cases are in the appendix. 
We can see that our MotionEditor can accomplish a wide range of motion editing while simultaneously preserving the original protagonist's appearance and background information. 

\begin{figure}[t!]
\begin{center}
\includegraphics[width=1\linewidth]{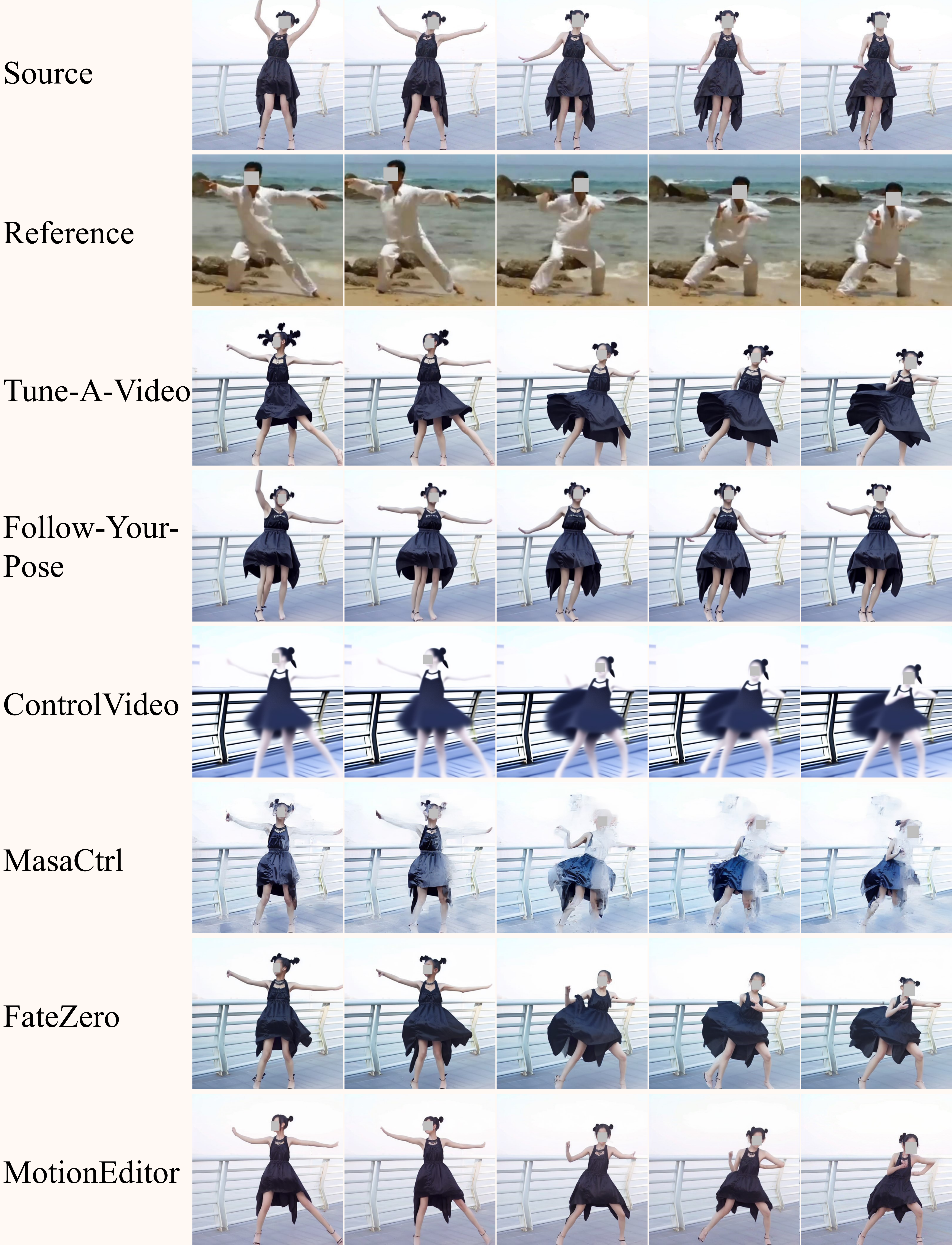}
\end{center}
\vspace{-0.3cm}
   \caption{Qualitative comparison between our MotionEditor and other state-of-the-art video editing models. 
   Source prompt: ``a girl in a black dress is dancing."
   Target prompt: ``a girl in a black dress is practicing tai chi."
   Our method exhibits accurate motion editing and appearance preservation.
   }
\label{fig:comparison_result}
\vspace{-0.2cm}
\end{figure}

\subsection{Comparison with State-of-the-Art Methods}
\textbf{Competitors.} We compare our MotionEditor against recent approaches to validate the superiority of our model. The competitors are depicted as follows: 
(1) Models based on GANs for human motion transfer, including LWG~\cite{liu2019liquid} that disentangles the pose and shape, and MRAA~\cite{siarohin2021motion} that learns semantic object parts.
(2) Tune-A-Video~\cite{wu2023tune}, which inflates Stable Diffusion in a one-shot learning manner. 
(3) Follow-Your-Pose~\cite{ma2023follow}, which proposes a spatial pose adapter for pose-guided video generation.
(4) ControlVideo~\cite{zhang2023controlvideo}, which designs fully cross-frame attention, attempting to stitch video frames into one large image. 
(5) MasaCtrl~\cite{cao2023masactrl}, which conducts mask-guided mutual self-attention fusion, in which the masks are derived from text cross attention. 
(6) FateZero~\cite{qi2023fatezero}, which designs inversion attention fusion for retaining source structure. 
It is worth noting that Tune-A-Video and Masactrl are equipped with a conditional T2I model (\emph{i.e.} ControlNet) for controllable video editing.
We also combine FateZero with ControlNet to enable pose-guided editing.
More details on implementation are depicted in the appendix.

\textbf{Qualitative results.} 
We conduct a qualitative comparison of our MotionEditor against several competitors in Fig. \ref{fig:comparison_result}. The results of LWG~\cite{liu2019liquid} and MRAA~\cite{siarohin2021motion} are provided in the appendix. By analyzing the results, we have the following observations: 
(1) Tune-A-Video~\cite{wu2023tune}, ControlVideo~\cite{zhang2023controlvideo}, Masactrl~\cite{cao2023masactrl} and FateZero~\cite{qi2023fatezero} demonstrate capabilities in editing motion to a certain extent.
Nevertheless, their edited videos exhibit a considerable degree of ghosting, manifested as overlapping frames of the protagonist's heads and legs. 
(2) Follow-Your-Pose~\cite{ma2023follow} fails to perform motion editing and preserve the source appearance. The ghosting effect also exists in the result. The plausible reason is that it has difficulties handling the motion conflict between inverted noise and input reference pose.
(3) Tune-A-Video~\cite{wu2023tune} and ControlVideo~\cite{zhang2023controlvideo} both have dramatic appearance changes, \emph{e.g.}, hairstyle, texture of clothes and background. 
Due to the constraint of a single-branch architecture, they are unable to interact with the original video features when additional pose conditions are introduced. 
This leads to a gradual loss of the source appearance along with the increasing denoising steps. 
(4) Masactrl~\cite{cao2023masactrl} generates an excessive amount of blurry noise. 
The possible reason is that the masks generated by cross attention maps are unreliable and inconsistent across time, thus introducing blurring noise to the result.
(5) The edited result of FateZero~\cite{qi2023fatezero} shows overlapping frames of the protagonist's legs. 
It demonstrates that the attention fusion strategy in FateZero may not be suitable for motion editing. 
(6) Finally, our MotionEditor can effectively perform motion editing while preserving the original background and appearance compared with previous approaches, highlighting its great potential.

\begin{table}[t!]\small
\caption{Quantitative comparisons on 20 in-the-wild cases. L-S, L-N, and L-T indicate LPIPS-S, LPIPS-N, LPIPS-T respectively. 
}
\vspace{-0.1in}
\begin{center}
\renewcommand\arraystretch{1.1}
\scalebox{0.9}{
\begin{tabular}{lcccc}
\toprule
Method           & CLIP ($\uparrow$)           & L-S ($\downarrow$)       & L-N ($\downarrow$)        & L-T ($\downarrow$)       \\ \midrule
LWG~\cite{liu2019liquid}              & 25.35          & 0.431          & 0.194          & 0.203          \\
MRAA~\cite{siarohin2021motion}             & 26.80          & 0.462          & 0.269          & 0.353          \\
Tune-A-Video~\cite{wu2023tune}     & 27.71          & 0.345          & 0.169          & 0.157          \\
Follow-Your-Pose~\cite{ma2023follow} & 26.55          & 0.337          & 0.144          & 0.183          \\
ControlVideo~\cite{zhang2023controlvideo}     & 26.87          & 0.428          & 0.228          & 0.311          \\
MasaCtrl~\cite{cao2023masactrl}         & 27.14          & 0.372          & 0.236          & 0.177          \\
FateZero~\cite{qi2023fatezero}         & 28.07          & 0.308          & 0.176          & 0.124          \\ \midrule
MotionEditor     & \textbf{28.86} & \textbf{0.273} & \textbf{0.124} & \textbf{0.082} \\ \bottomrule
\end{tabular}
}
\end{center}
\label{table:quantitative_comparisons}
\vspace{-0.2in}
\end{table}

\textbf{Quantitative results.} To the best of our knowledge, there is still no widely recognized metric for evaluating the performance of video editing. 
In this paper, we conduct quantitative comparisons against previous approaches through several perceptual metrics~\cite{zhang2018perceptual} and a user study on edited videos.
The detailed metrics are as follows: (1) CLIP score~\cite{radford2021learning}: Target textual faithfulness. (2) LPIPS-S: Learned Perceptual Image Patch Similarity (LPIPS)~\cite{zhang2018perceptual} between edited frames and source frames. (3) LPIPS-N: LPIPS between edited neighboring frames. 
(4) LPIPS-T: We split a long video into two segments. The first segment is used for the source, while the second segment is for the reference. We compute LPIPS between the edited video and the second segment. The results are shown in Table \ref{table:quantitative_comparisons}.
We observe that MotionEditor surpasses the competitors by a large margin.

\begin{table}[t!]\small
\caption{User preference ratio of MotionEditor when comparing with each method. Higher indicates the users prefer more to our MotionEditor. M-A, A-A and T-A indicate motion alignment, appearance alignment, and textual alignment, respectively.}
\vspace{-0.1in}
\begin{center}
\renewcommand\arraystretch{1.1}
\scalebox{0.95}{
\begin{tabular}{lccc}
\toprule
Method           & M-A & A-A & T-A \\ \midrule
LWG~\cite{liu2019liquid}              & 91.9\%       & 95.7\%           & 90.0\%       \\
MRAA~\cite{siarohin2021motion}             & 94.8\%       & 98.4\%           & 92.6\%        \\
Tune-A-Video~\cite{wu2023tune}     & 87.6\%      & 90.8\%           & 79.2\%       \\
Follow-Your-Pose~\cite{ma2023follow} & 96.3\%       & 85.0\%          & 84.6\%       \\
ControlVideo~\cite{zhang2023controlvideo}     & 94.1\%       & 98.8\%           & 86.0\%       \\
MasaCtrl~\cite{cao2023masactrl}         & 89.4\%      & 94.6\%           & 87.8\%       \\
FateZero~\cite{qi2023fatezero}         & 78.9\%      & 74.5\%          & 74.7\%      \\ \bottomrule
\end{tabular}
}
\end{center}
\label{table:user_study}
\vspace{-0.2in}
\end{table}

We also conduct a user study to evaluate the human preference between our method and the competitors.
For each case, participants are first presented with the source and target videos, as well as the prompts.
We then show two motion-edited videos; one is generated by our method and the other is from a competitor, in random order.
Participants are asked to answer the following questions: ``which one has better motion alignment with reference", 
``which one has better appearance alignment with source", and
``which one has better content alignment with the prompt."
The total number of cases is 20, and the participants are mainly university students.
The results in Table~\ref{table:user_study} show that our method is far ahead of other methods in terms of subjective evaluation.

\begin{figure}[t!]
\begin{center}
\includegraphics[width=1\linewidth]{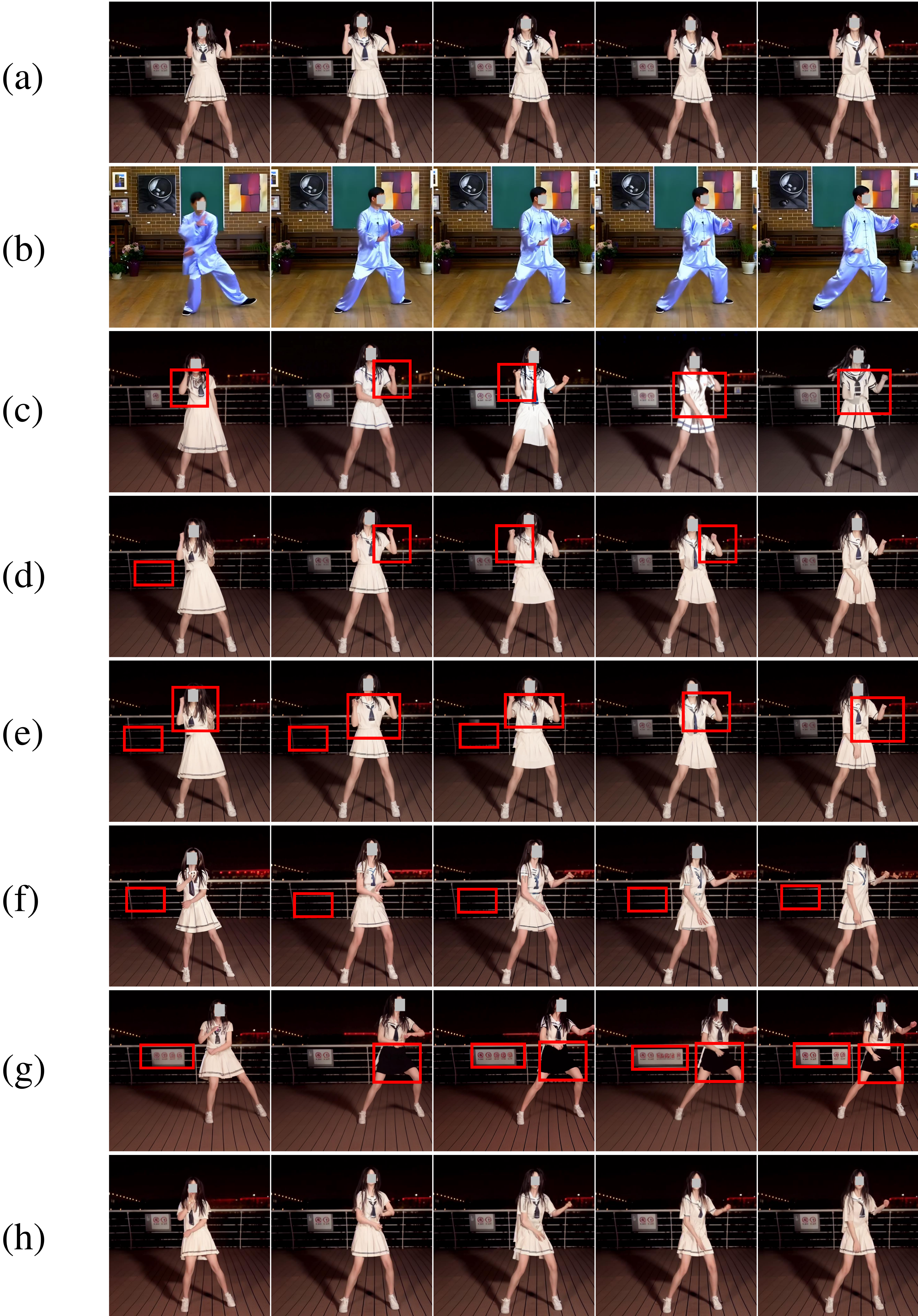}
\end{center}
   \caption{Ablations on core components of MotionEditor.
   Rows in the figure are: (a) source, (b) reference, (c) \textit{w/o} CS Attention, (d) \textit{w/o} cross attention in motion adapter, (e) \textit{w/o} motion adapter, (f) \textit{w/o} high-fidelity attention injection, (g) \textit{w/o} skeleton alignment, and (h) MotionEditor.
   Source prompt: ``A girl is dancing." Target prompt: ``A girl is practicing tai chi."}
\label{fig:ablation}
\end{figure}

\subsection{Ablation Study}
To validate the importance of the core components in MotionEditor, we conduct an ablation study. The results are illustrated in Fig. \ref{fig:ablation}. It is worth noting that we replace our proposed CS Attention with the previous Sparse Attention~\cite{wu2023tune} in (c). 
The results in row (c) indicate that Sparse Attention attempts to force the frames to be aligned with the first frame, resulting in unreliable motion. 
Rows (d) and (e) both fail to accomplish motion editing and background preservation. It demonstrates that the original ControlNet has weak constraints on the motion without additional content-aware modeling. It also forces the model to retain background information.
In row (f), model \textit{w/o} high-fidelity attention injection loses the original background details as the road sign behind the girl has disappeared. It validates that our proposed mechanism can promote the model to preserve the source background. 
Model \textit{w/o} skeleton alignment suffers from appearance change due to the misalignment, as in row (g). 
The misalignment of the skeletons may potentially introduce unexpected noise to the content latent, thus destroying source data distribution. The above ablation results demonstrate that our core components certainly contribute to the promising motion editing capability of MotionEditor.

\section{Conclusion}
\label{sec:conclusion} In this paper, we proposed MotionEditor for tackling video motion editing challenges, which is rated as high-level video editing compared with previous video attribute editing. 
To enhance the motion controllability, a content-aware motion adapter was designed to build up a relationship with the source content, enabling seamless motion editing as well as temporal modeling.
We further proposed a high-fidelity attention injection for preserving the source appearance of the background and protagonist. 
To alleviate the misalignment problem of skeleton signals, we presented a simple yet effective skeleton alignment to normalize the target skeletons. 
In conclusion, MotionEditor explores the rarely studied video motion editing task, encouraging more future studies in this challenging scenario.

{
    \small

}

\newpage

\appendix

\section*{Appendix}

\section{Skeleton Signal Alignment}
\label{sec:alignment}

The details of our skeleton signal alignment are depicted in Alignment \ref{alg:alignment}. $({x}_{s}, {y}_{s})$, $h_{s}$, $w_{s}$ refer to the coordinates of the upper left corner, the height, and the width for the bounding box of source protagonist.
$({x}_{r}, {y}_{r})$, $h_{r}$, $w_{r}$ refer to the counterparts in reference frame, \emph{i.e.}, the coordinates of the upper left corner, the height, and the width for the bounding box of reference protagonist.
${w}_{r}^{*}$ indicates the resized width of source protagonist. 
$\mathtt{Rectangle}\_\mathtt{Boundary}(\cdot)$ indicates $cv2.boundingRect(\cdot)$.

\begin{algorithm}[h]
\caption{Skeleton Alignment}
\label{alg:alignment}
\begin{algorithmic}
\small 
\State \textbf{Input:} Source Skeleton $\bm{S}_{sr}$, Source Mask $\bm{M}_{sr}$; Reference Skeleton $\bm{S}_{rf}$, Reference Mask $\bm{M}_{rf}$ \\
\\

 $\triangleright$ Resize Operation 
 
 \State $({x}_{s}, {y}_{s}), {h}_{s}, {w}_{s}=\mathtt{Rectangle}\_\mathtt{Boundary}(\bm{M}_{sr})$ 
 \State $({x}_{r}, {y}_{r}), {h}_{r}, {w}_{r}=\mathtt{Rectangle}\_\mathtt{Boundary}(\bm{M}_{rf})$ 
 \State $\bm{ratio}={w}_{r} / \mathtt{float}({h}_{r})$
 \State ${w}_{r}^{*}=\mathtt{Round}(\bm{ratio} \cdot {h}_{s})$
 \State $\bm{P}_{S}=\mathtt{resize}(\bm{S}_{rf}[~{y}_{r}:{y}_{r}+{h}_{r},{x}_{r}:{x}_{r}+{w}_{r}~], ({h}_{s}, {w}_{r}^{*}))$
 \State $\bm{P}_{M}=\mathtt{resize}(\bm{M}_{rf}[~{y}_{r}:{y}_{r}+{h}_{r},{x}_{r}:{x}_{r}+{w}_{r}~], ({h}_{s}, {w}_{r}^{*}))$
 \State $\bm{S}_{rf}=\mathtt{zeros} \textunderscore \mathtt{like}(\bm{S}_{sr})$ 
 \State $\bm{M}_{rf}=\mathtt{zeros} \textunderscore \mathtt{like}(\bm{M}_{sr})$
 \State if ${w}_{r}^{*} < {w}_{s}:$
 \State \hspace{1em} $\bm{S}_{rf}[~{y}_{s}:{y}_{s}+{h}_{s},{x}_{s}:{x}_{s}+{w}_{r}^{*}~]=\bm{P}_{S}$ 
 \State \hspace{1em} $\bm{M}_{rf}[~{y}_{s}:{y}_{s}+{h}_{s},{x}_{s}:{x}_{s}+{w}_{r}^{*}~]=\bm{P}_{M}$
 \State else$:$
 \State \hspace{1em} $\bm{S}_{rf}[~{y}_{s}:{y}_{s}+{h}_{s},{x}_{s}-({w}_{r}^{*}-{w}_{s}):{x}_{s}+{w}_{s}~]=\bm{P}_{S}$ 
 \State \hspace{1em} $\bm{M}_{rf}[~{y}_{s}:{y}_{s}+{h}_{s},{x}_{s}-({w}_{r}^{*}-{w}_{s}):{x}_{s}+{w}_{s}~]=\bm{P}_{M}$
 
\\
\State $\triangleright$ Translation Operation
\State $\mathbf{coordinates}_{s}=\mathtt{where}(\bm{M}_{sr}==1)$
\State $\mathbf{coordinates}_{r}=\mathtt{where}(\bm{M}_{rf}==1)$
\State $\mathbf{center}_{s}=\mathtt{Mean}(\mathbf{coordinates}_{s}, \mathtt{axis}=0)$
\State $\mathbf{center}_{r}=\mathtt{Mean}(\mathbf{coordinates}_{r}, \mathtt{axis}=0)$
\State $\bm{v}_{trans}=\mathbf{center}_{s}-\mathbf{center}_{r}$
\State ${dx}=\bm{v}_{trans}[~0~],~~{dy}=\bm{v}_{trans}[~1~]$
\State $\bm{M}_{trans}=[[~1, 0, {dx}~],~[~0, 1, {dy}~]]$
\State $\bar{\bm{S}}_{tg}=\mathtt{WarpAffine}(\bm{S}_{rf}, \bm{M}_{trans}, \bm{S}_{rf}.shape)$
\State {\bf Output:} Aligned Target Skeleton $\bar{\bm{S}}_{tg}$  
\end{algorithmic}
\end{algorithm}

\section{Tasks Comparison}
\label{sec:tasks}

We propose MotionEditor to tackle a higher-level and more challenging video editing---video motion editing. Given the source video, target prompt, and reference video, our model can directly edit the motion of the source video according to that of the reference video and the description of the target, while preserving the appearance information of the source video. The comparison details are depicted in Table \ref{table:tasks}. 

\begin{table*}[t!]\small
\caption{Tasks comparison among pose-guided image generation, human motion transfer, pose-guided video generation, video attribute editing, and video motion editing.  
}
\vspace{-0.1in}
\begin{center}
\renewcommand\arraystretch{1.1}
\scalebox{0.9}{
\begin{tabular}{lcc}
\hline
\textbf{Task}                & \textbf{Input}                      & \textbf{Output} \\ \toprule
Pose-guided Image Generation~(ControlNet~\cite{zhang2023adding}) & Prompt~+~Pose                         & Image           \\
Human Motion Transfer~(LWG~\cite{liu2019liquid} and MRAA~\cite{siarohin2021motion})        & Image~+~Series of Poses               & Video           \\
Pose-guided Video Generation~(Follow-Your-Pose~\cite{ma2023follow} and ControlVideo~\cite{zhang2023controlvideo}) & Prompt~+~Series of Poses              & Video           \\
Video Attribute Editing~(Tune-A-Video~\cite{wu2023tune}, MasaCtrl~\cite{cao2023masactrl} and FateZero~\cite{qi2023fatezero})      & Video~+~Prompt                        & Video           \\ \midrule
Video Motion Editing~(MotionEditor)         & Video~+~Series of Poses~+~Prompt & Video           \\ \bottomrule
\end{tabular}
}
\end{center}
\label{table:tasks}
\vspace{-0.2in}
\end{table*}

\section{Experiment Details}
\label{sec:details}

Since existing methods are not designed for motion editing, we make several modifications to their models.
For pose-guided video generation models, we input DDIM inverted source video latent to Follow-Your-Pose~\cite{ma2023follow} and ControlVideo~\cite{zhang2023controlvideo}, which enables them to perform controllable video editing. 
For video attribute editing models, Tune-A-Video~\cite{wu2023tune}, MasaCtrl~\cite{cao2023masactrl} and FateZero~\cite{qi2023fatezero} are equipped with ControlNet~\cite{zhang2023adding} which enables them to accept additional controllable signals inputs. 
In terms of human motion transfer models, we only feed the first frame of each video to LWG~\cite{liu2019liquid} and MRAA~\cite{siarohin2021motion} to follow their original pipelines. The remaining settings are identical to those in our proposed MotionEditor. 
We illustrate the video motion editing results of 32 frames with a $4\times$ sampling ratio for better visual appeal.

\begin{figure}
\begin{center}
\includegraphics[width=1\linewidth]{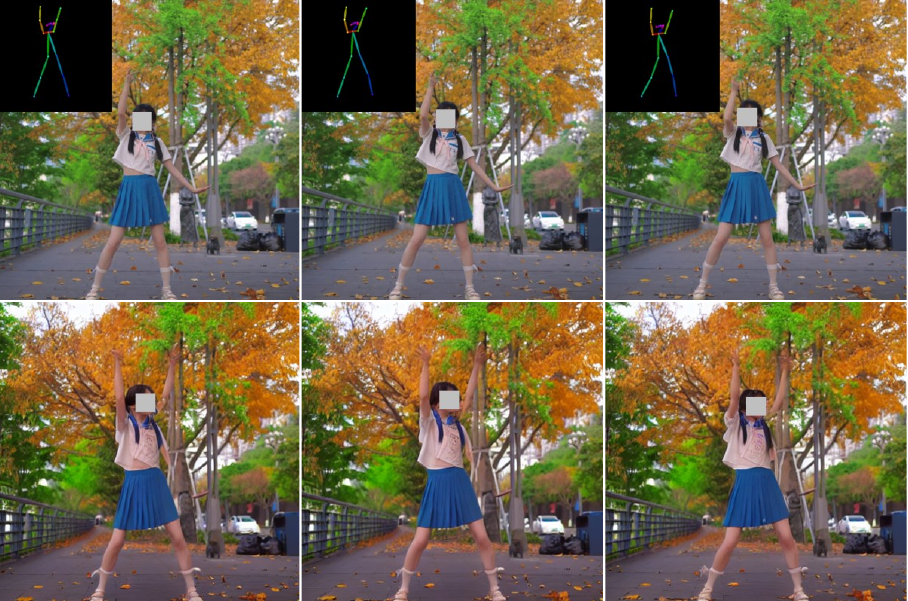}
\end{center}
\vspace{-0.5cm}
   \caption{The failure case of our MotionEditor.
   }
\label{fig:failure}
\vspace{-0.5cm}
\end{figure}

\section{Discussion on Video Motion Editing and Human Motion Transfer}
\label{sec:discussion}

\begin{figure*}[t!]
\begin{center}
\includegraphics[width=1\linewidth]{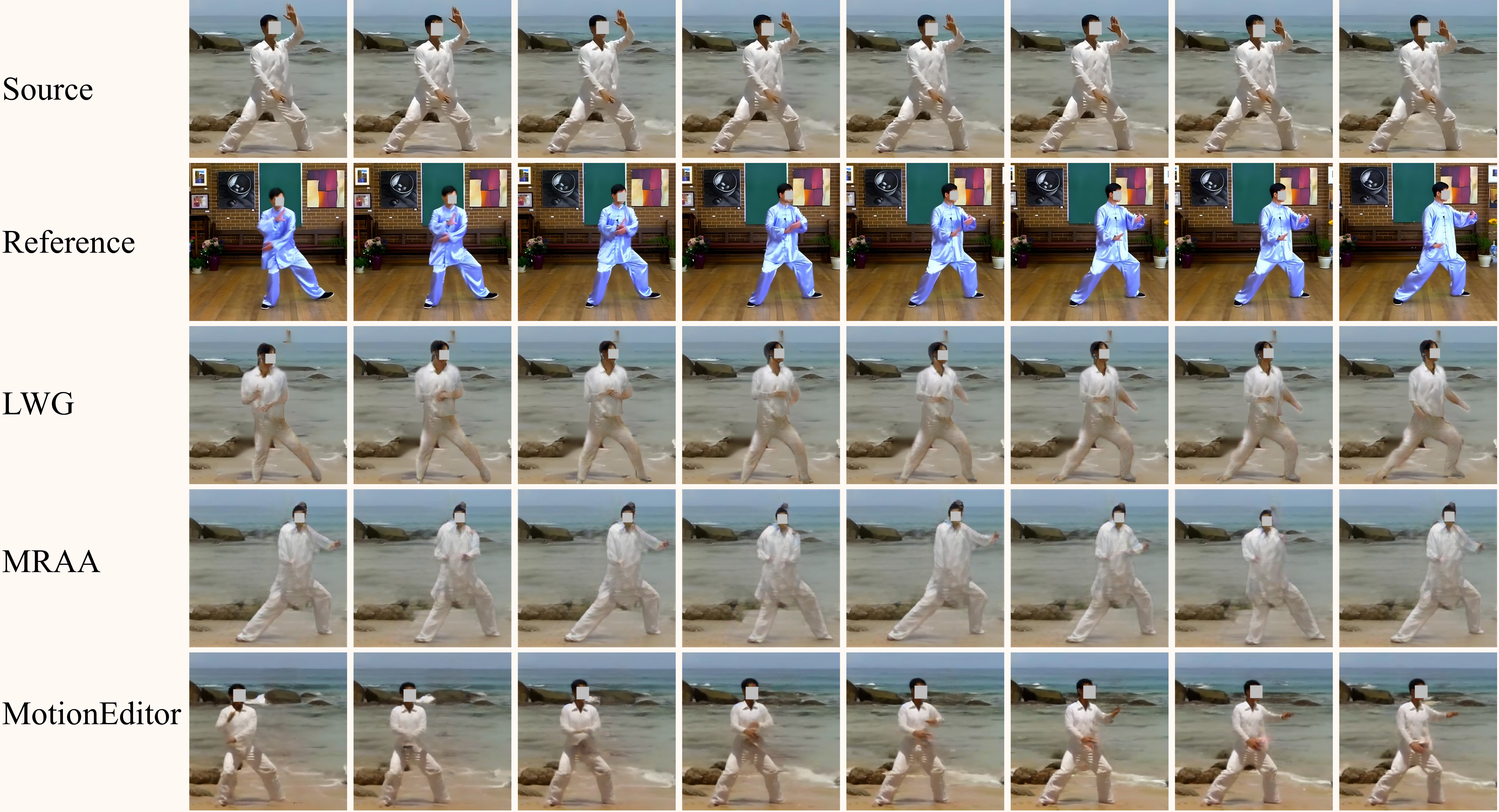}
\end{center}
\vspace{-0.5cm}
   \caption{Comparison between video motion editing and human motion transfer with dynamic backgrounds.
   }
\label{fig:discussion_background}
\vspace{-0.1cm}
\end{figure*}

\begin{figure*}[t!]
\begin{center}
\includegraphics[width=1\linewidth]{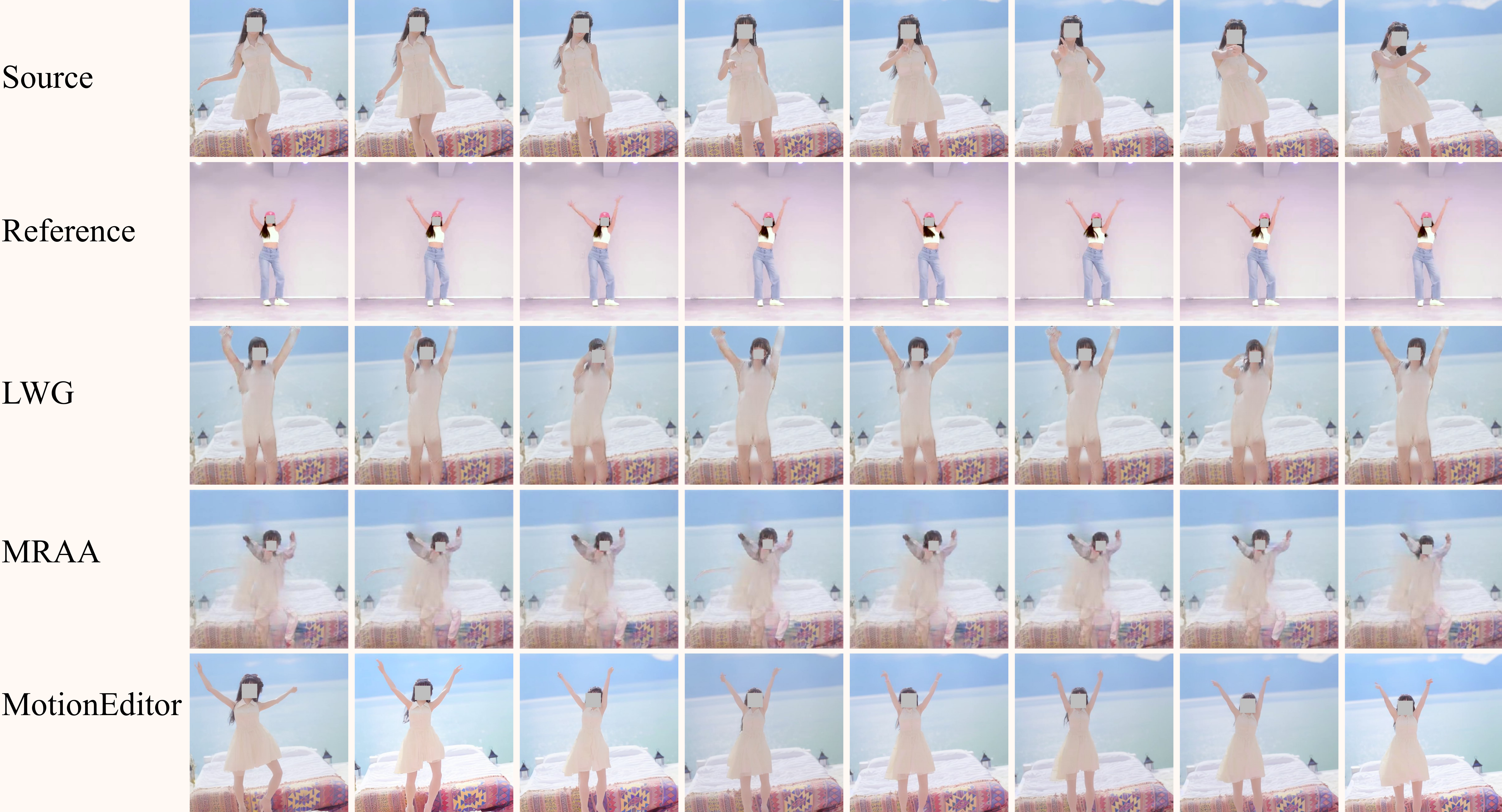}
\end{center}
\vspace{-0.5cm}
   \caption{Comparison between video motion editing and human motion transfer with camera movement.
   }
\label{fig:discussion_camera}
\vspace{-0.1cm}
\end{figure*}

Video Motion Editing requires directly performing motion transfer on video over temporal dimension, which considers the per-frame dynamic background information and camera movement. In contrast, the pipeline of human motion transfer~\cite{liu2019liquid,siarohin2021motion} only demands one single image which ignores the additional dynamic information. We conduct two comparison experiments on a case with dynamic background information and a case with camera movement. The results are shown in Figure \ref{fig:discussion_background} and Figure \ref{fig:discussion_camera}. We can see that our MotionEditor can perform motion editing in higher quality while maintaining the additional dynamic information. The results of human motion transfer models are limited to a single given image, thereby resulting in static background information and camera movement. In addition, they are usually constrained to images with clean backgrounds. When the background is intricate with complex scenes, their models exhibit limited capabilities.

\begin{figure*}[t!]
\begin{center}
\includegraphics[width=1\linewidth]{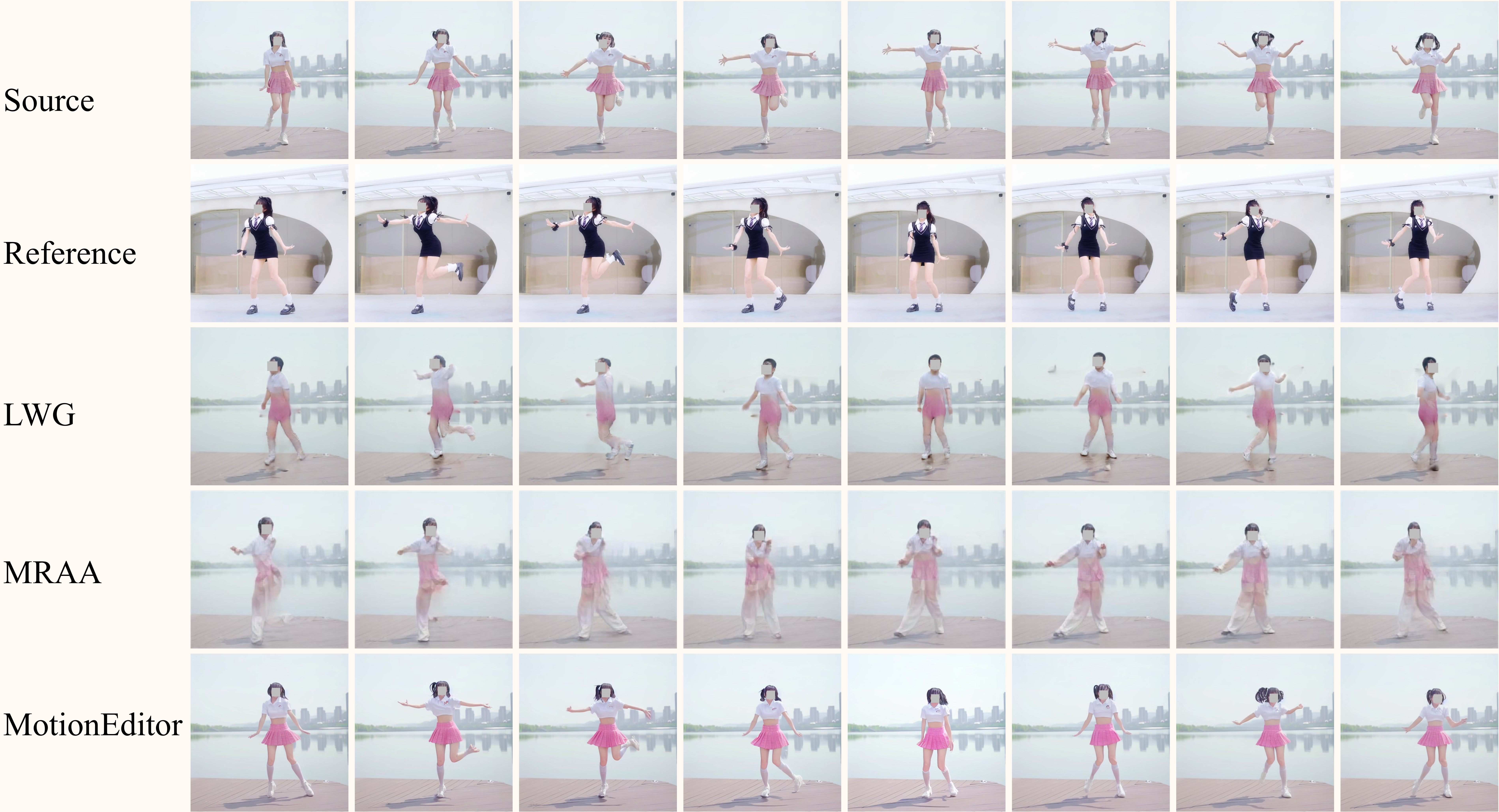}
\end{center}
\vspace{-0.5cm}
   \caption{Comparison between video motion editing and human motion transfer with complex initial poses.
   }
\label{fig:discussion_complex}
\vspace{-0.1cm}
\end{figure*}

Human motion transfer methods (\emph{e.g.}, LWG~\cite{liu2019liquid} and MRAA~\cite{siarohin2021motion}) are also sensitive to the initial source pose.
They demand simple initial source poses, for example, a person standing squarely.
We conduct an experiment that selects frames with different and complex poses as the initial image for motion transfer.
The result is illustrated in Figure \ref{fig:discussion_complex}. 
Note that only the first frame of each transfer is presented.
It indicates that human motion transfer models fail to directly handle different and complex initial poses while our MotionEditor shows its superiority in video motion editing.

\section{Additional Results}
\label{sec:additional_results}

\begin{figure*}[t!]
\begin{center}
\includegraphics[width=1\linewidth]{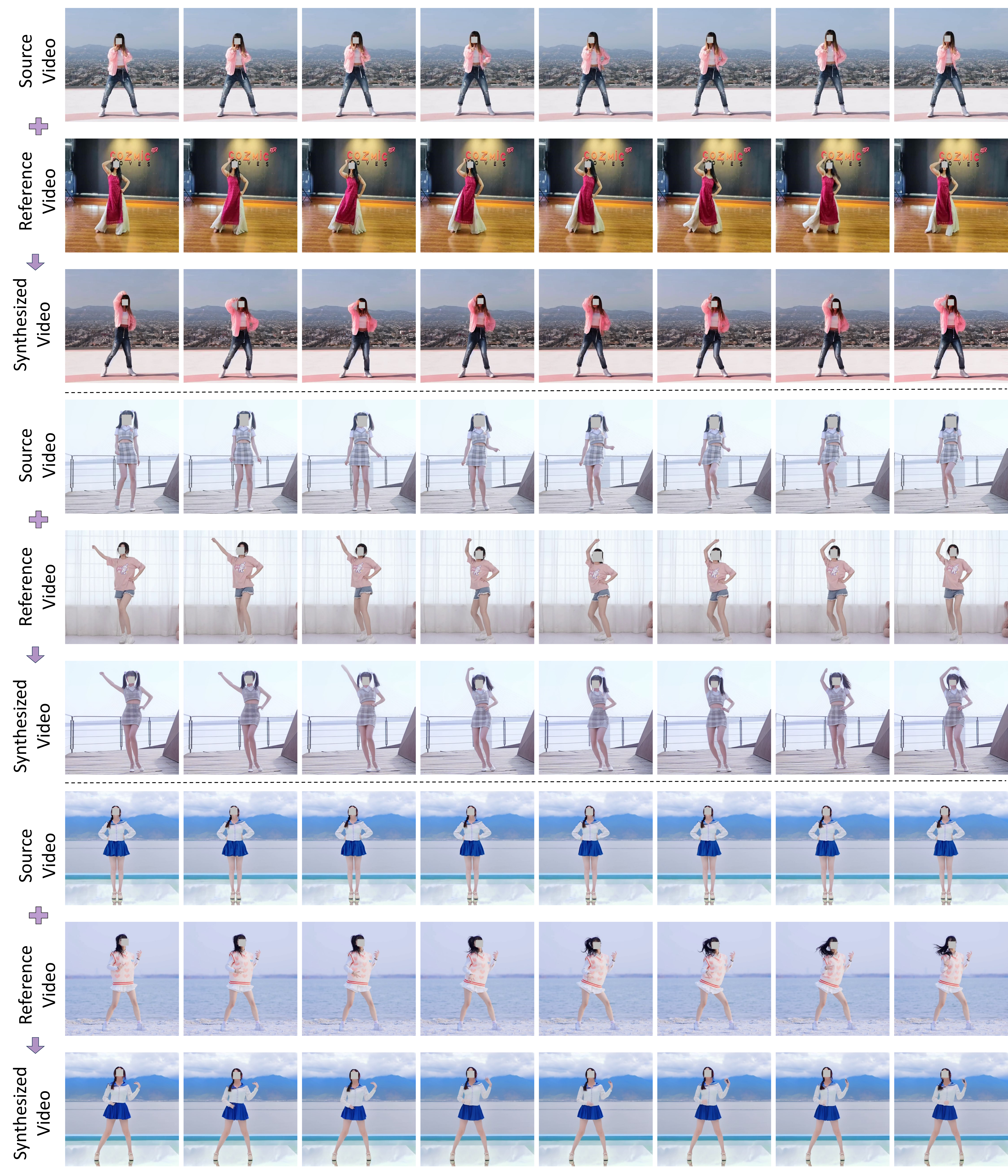}
\end{center}
\vspace{-0.5cm}
   \caption{Additional video motion editing results (1/2).
   }
\label{fig:additional_result_1}
\vspace{-0.1cm}
\end{figure*}

\begin{figure*}[t!]
\begin{center}
\includegraphics[width=1\linewidth]{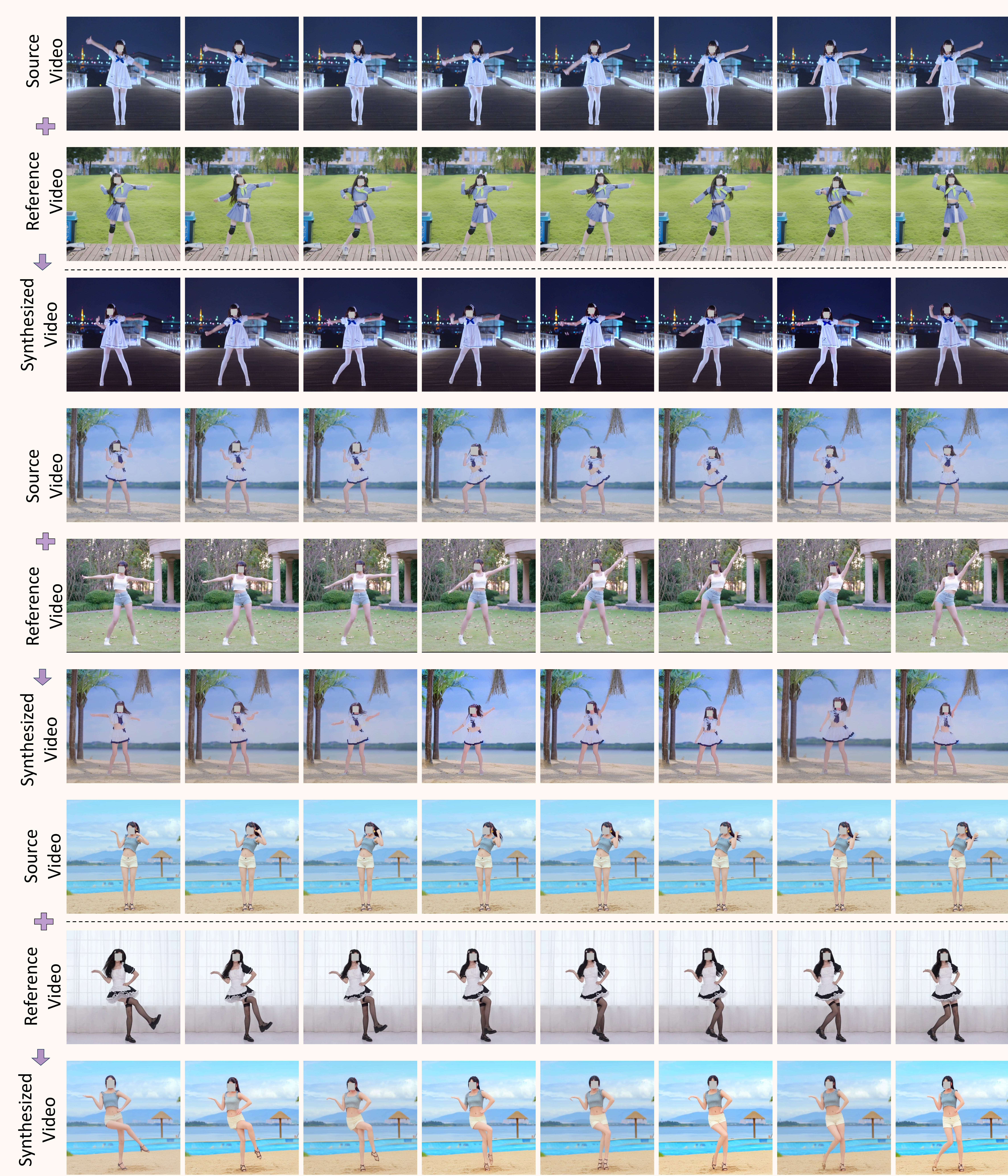}
\end{center}
\vspace{-0.5cm}
   \caption{Additional video motion editing results (2/2).
   }
\label{fig:additional_result_2}
\vspace{-0.1cm}
\end{figure*}

\begin{figure*}[t!]
\begin{center}
\includegraphics[width=1\linewidth]{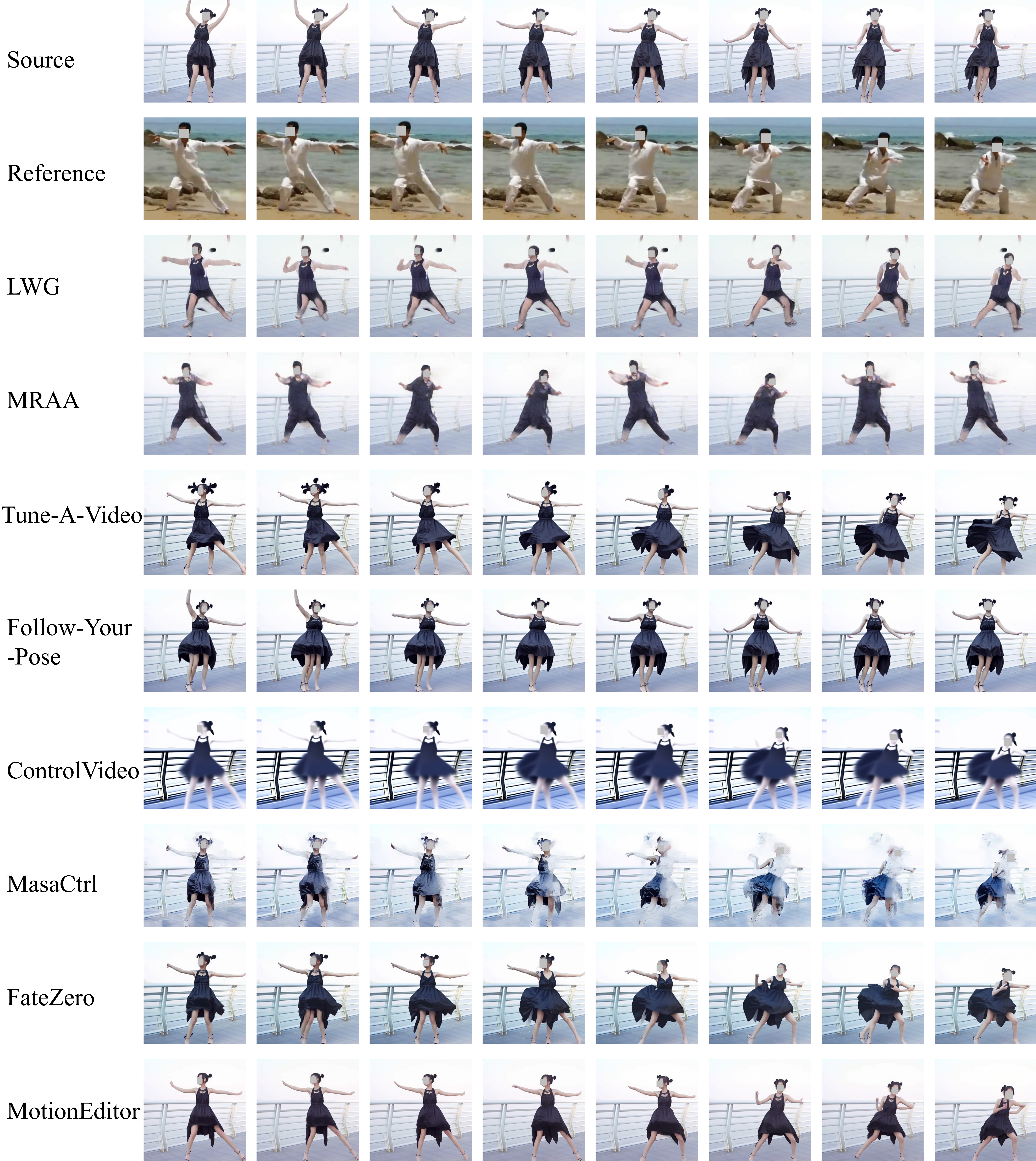}
\end{center}
\vspace{-0.5cm}
   \caption{Complete video motion editing comparison results.
   }
\label{fig:comparison_result_1}
\vspace{-0.1cm}
\end{figure*}

\begin{figure*}[t!]
\begin{center}
\includegraphics[width=1\linewidth]{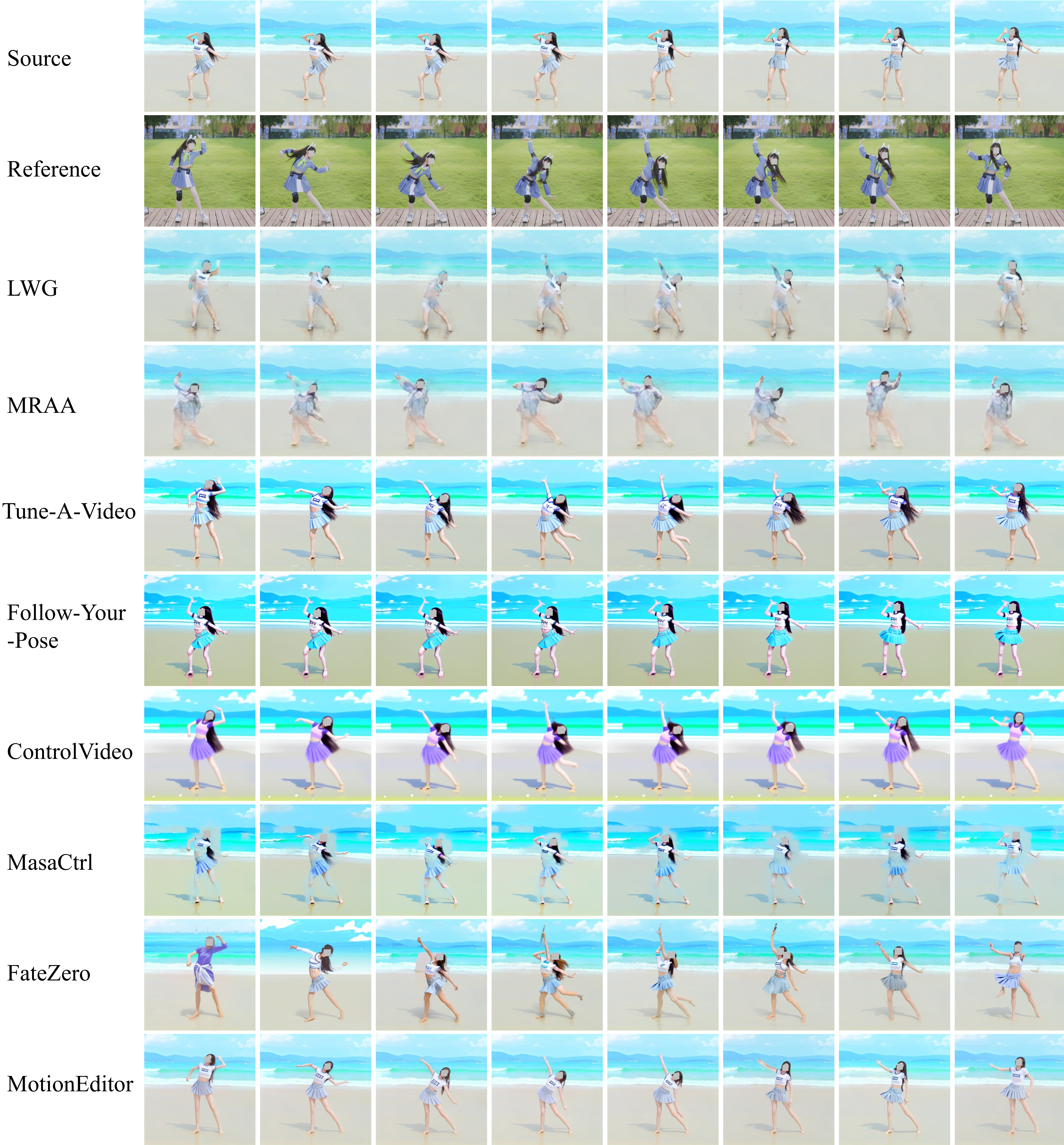}
\end{center}
\vspace{-0.5cm}
   \caption{Additional video motion editing comparison results.
   }
\label{fig:additional_comparison_result}
\vspace{-0.1cm}
\end{figure*}

\begin{figure*}[t!]
\begin{center}
\includegraphics[width=1\linewidth]{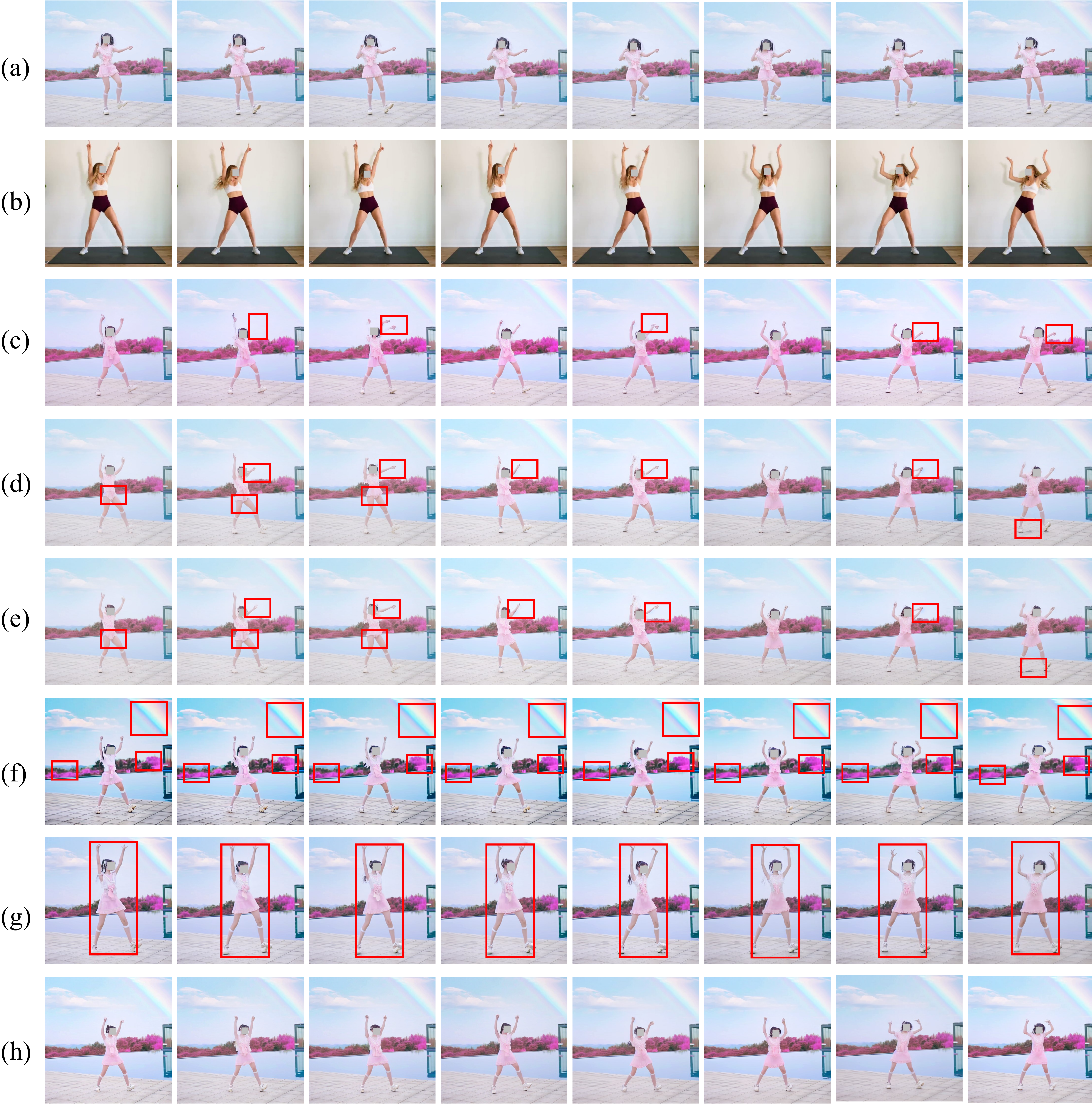}
\end{center}
\vspace{-0.5cm}
   \caption{Additional ablation study results. Rows in the figure are: (a) source, (b) reference, (c) \textit{w/o} CS Attention, (d) \textit{w/o} cross attention in motion adapter, (e) \textit{w/o} motion adapter, (f) \textit{w/o} high-fidelity attention injection, (g) \textit{w/o} skeleton alignment, and (h) MotionEditor.
   }
\label{fig:additional_ablation}
\vspace{-0.1cm}
\end{figure*}

Figure \ref{fig:additional_result_1} and \ref{fig:additional_result_2} show additional video motion editing results of MotionEditor. Figure \ref{fig:comparison_result_1} shows the complete comparison results. Figure \ref{fig:additional_comparison_result} and \ref{fig:additional_ablation} provide additional comparison results and ablation study results.

\section{Limitations and Future Work}
\label{sec:limitation}

Figure \ref{fig:failure} shows one failure case of our MotionEditor. The hands of the girl are confused with the surrounding background. The plausible reason is that the foreground latents are confused with background latents, thereby introducing additional bias to self-attention in the U-Net during the denoising process. The probable solution is to explicitly decouple the foreground and background before the denoising process and perform the corresponding editing operation on them. Moreover, a learnable dedicated mixture adapter can be designed for blending foreground with background naturally. This part is left as future work.

\end{document}